\documentclass[10pt,twocolumn,letterpaper]{article}

\usepackage{iccv}
\usepackage{times}
\usepackage{epsfig}
\usepackage{graphicx}
\usepackage{amsmath}
\usepackage{amssymb}
\usepackage{subfigure}
\usepackage{tabularx}
\usepackage{booktabs}
\usepackage{multirow}
\usepackage{float}
\usepackage{bm}
\usepackage{caption}
\usepackage{comment}
\usepackage[ruled,vlined]{algorithm2e}
\usepackage[numbers,sort&compress]{natbib}
\def\etal{\emph{et al.~}}

\newcommand\blfootnote[1]{%
  \begingroup
  \renewcommand\thefootnote{}\footnote{#1}%
  \addtocounter{footnote}{-1}%
  \endgroup
}

\usepackage[pagebackref=true,breaklinks=true,letterpaper=true,colorlinks,bookmarks=false]{hyperref}

\iccvfinalcopy 

\ificcvfinal\fi

\begin{document}

\title{Everybody Is Unique: Towards Unbiased Human Mesh Recovery}

\author{Ren Li, Meng Zheng, Srikrishna Karanam$^{*}$, Terrence Chen, and Ziyan Wu\\
United Imaging Intelligence, Cambridge MA\\
{\tt \{first.last\}@uii-ai.com}
}


\twocolumn[{%
\renewcommand\twocolumn[1][]{#1}%
\maketitle
\begin{center}
    \centering
    \includegraphics[width=1.0\linewidth]{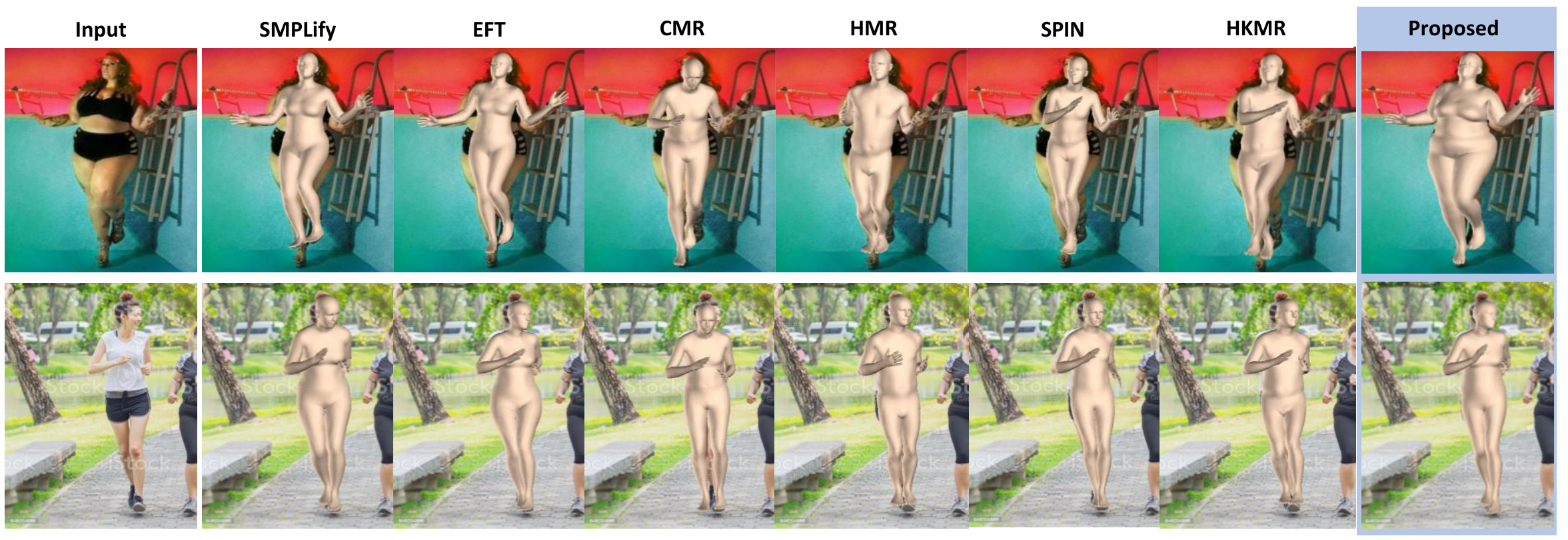}
    \captionof{figure}{We propose a generalized human mesh optimization algorithm that produces substantially improved results without needing 3D mesh annotations across both standard benchmark data and application-specific data such as obese person images.}
    \label{fig:teaser}
\end{center}%
}]

\ificcvfinal\thispagestyle{empty}\fi

\begin{abstract}

We consider the problem of obese human mesh recovery, i.e., fitting a parametric human mesh to images of obese people. Despite obese person mesh fitting being an important problem with numerous applications (e.g., healthcare), much recent progress in mesh recovery has been restricted to images of non-obese people. In this work, we identify this crucial gap in the current literature by presenting and discussing limitations of existing algorithms. Next, we present a simple baseline to address this problem that is scalable and can be easily used in conjunction with existing algorithms to improve their performance. Finally, we present a generalized human mesh optimization algorithm that substantially improves the performance of existing methods on both obese person images as well as community-standard benchmark datasets. A key innovation of this technique is that it does not rely on supervision from expensive-to-create mesh parameters. Instead, starting from widely and cheaply available 2D keypoints annotations, our method automatically generates mesh parameters that can in turn be used to re-train and fine-tune any existing mesh estimation algorithm. This way, we show our method acts as a drop-in to improve the performance of a wide variety of contemporary mesh estimation methods. We conduct extensive experiments on multiple datasets comprising both standard and obese person images and demonstrate the efficacy of our proposed techniques.

\end{abstract}

\section{Introduction}
\label{sec:intro}
\blfootnote{* This work was done during Ren Li's internship with United Imaging Intelligence. Corresponding author: Srikrishna Karanam.}We consider the problem of human mesh estimation. Given a person image and a functionally-known parametric human mesh, the problem is to fit the mesh (i.e., estimate its parameters) so as to best explain the 3D pose and shape of the person. With many important real-world applications \cite{karanam2020towards,ching2014patient}, there has been much recent progress in this field as evidenced by the dramatic reduction in the mean per joint position error (MPJPE) metric on benchmark datasets \cite{kanazawa2018end, kolotouros2019learning, georgakis2020hierarchical}. In particular, given the ongoing COVID-19 pandemic, there has been a substantial increase in the need for healthcare applications such as patient positioning \cite{karanam2020towards} for CT/MR scans \cite{fang2020sensitivity}, leading to an immediate practical relevance for human mesh estimation/modeling. 

Applications such as the above demand systems that are robust on a wide variety of data. One such dimension of diversity is the physical size of the person of interest. With obesity prevailing in about 40\% of the U.S. population \cite{kass2020obesity} and the medical devices industry consequently recognizing and developing scanners to particularly serve this community \cite{uMROmega}, it is critical that the underlying mesh estimation algorithms work well on images of such people. However, this problem has not attracted much attention in the mesh recovery research community, with Fig.~\ref{fig:teaser} showing some unsatisfactory results with the current state of the art, leading to biased estimates for obese person images.

In this paper, we identify this crucial gap in the current literature, and present and discuss the problem of obese human mesh recovery. We discuss how the current state-of-the-art methods fail to fit parametric meshes (e.g., SMPL \cite{loper2015smpl}, see Fig.~\ref{fig:teaser}) for obese data while working relatively reasonably well on standard benchmark data, leading to the issue of biased estimators. Among other reasons further discussed below, a key reason for this bias is lack of representative samples in current benchmark datasets. In the context of mesh estimation, in addition to insufficient raw data, this also manifests in the form of scarce annotations. 

As noted in prior work \cite{kanazawa2018end}, training reliable convolutional neural network (CNN) models requires large amounts of data annotated with mesh parameters. However, obtaining these annotations for non-obese person images is not trivial, let alone obese person images (e.g., for SMPL \cite{loper2015smpl}, this is an elaborate process involving ground-truth MoCap data and custom marker-based algorithms like MoSH \cite{loper2014mosh}). On the other hand, obtaining 2D annotations, e.g., image keypoints, is relatively straightforward and inexpensive as this can be accomplished using crowdsourcing platforms, e.g., Mechanical Turk, where requirements for technical know-how is limited (compared to MoCap algorithms like the above). Consequently, while datasets with 2D keypoint annotations can be found in abundance \cite{johnson2010clustered,johnson2011learning,andriluka20142d,lin2014microsoft}, those with full mesh annotations are substantially fewer \cite{ionescu2013human3}, with obese mesh annotations even more difficult to obtain. Given these considerations, we ask two key questions- (a) \textit{given abundant 2D keypoint annotations, can we automatically generate mesh parameters?}, and (b) \textit{can we develop an algorithm that is flexible to address the question above for both standard/non-obese person images in general and obese person images in particular?}

There has been some recent work \cite{bogo2016keep, joo2020exemplar} that propose optimization-based strategies for generating mesh parameters from 2D keypoints. However, a number of issues preclude their use for both general as well as obese mesh fitting. First, the work of Bogo \etal \cite{bogo2016keep} does not use the full context information provided by an input image, instead optimizing only a 2D-keypoint-based reprojection error objective, leading to the classic issue of depth ambiguity where multiple 3D configurations may correspond to the same 2D projection. While 3D pose and shape priors may address this issue to a certain extent, these constraints can only ensure the resulting fits belong to a pre-defined distribution. Furthermore, the current community-standard priors used for this purpose \cite{bogo2016keep, kanazawa2018end}, however, are not sufficiently representative of obese person images, leading to the unsatisfactory results of Fig.~\ref{fig:teaser} discussed earlier. Next, while the work of Joo \etal \cite{joo2020exemplar} addresses some of the aforementioned issues by optimizing the reprojection error cost function for the CNN model parameters (instead of SMPL mesh parameters as in Bogo \etal \cite{bogo2016keep}), its performance is also impacted by the priors issue noted above (see Fig.~\ref{fig:teaser} for results) since it starts the optimization from a CNN model that has been pre-trained on the same kind of data. Furthermore, this method has a trade-off between performance and number of optimization steps, increasing which can result in overfitting the reprojection objective. In fact, this is particularly
pronounced due to the lack of any regularizers to ensure anthropometric realism of the predicted mesh.

We take a structured approach to address the questions and issues noted above. First, we present a simple baseline approach that focuses on improving shape fits for obese person images. We achieve this by proposing a loss term that penalizes incorrect shape predictions by means of explicit 2D shape constraints. We show the proposed loss term can be flexibly and individually used in conjunction with both mesh- and CNN-parameter optimization strategies discussed above, thus immediately improving their performance (see Fig.~\ref{fig:teaser}). Next, we propose a generalization of this baseline that inherits the benefits of each strategy as part of an alternating directions scheme that optimizes for \textit{both} mesh- and CNN-parameters jointly (instead of separately as above). Our key insight is that such an alternating iterative framework leads to a virtuous cycle where the limitations
discussed above can be addressed in a principled manner. Specifically, the issue of depth ambiguity with Bogo \etal \cite{bogo2016keep} can be addressed with a pre-trained data-driven CNN model like in Joo \etal \cite{joo2020exemplar}, whereas the overfitting
problem of Joo \etal \cite{joo2020exemplar} can be addressed by using the
pose and shape fits generated by Bogo \etal \cite{bogo2016keep} as an explicit regularization term. Our baseline shape constraints can then be optionally added to the resulting overall learning objective, leading to a technique for both standard/non-obese as well as specific obese mesh fitting problems. 

We conduct numerous experiments to evaluate our proposed techniques. First, we discuss limitations of the standard MPJPE metric in capturing shape deviations (from ground truth) in obese images, leading to our new per-vertex error metric. Next, we show that the mesh fits produced by our shape-constraint baseline as well as its generalization outperform both Bogo \etal \cite{bogo2016keep} and Joo \etal \cite{joo2020exemplar} individually on both obese as well as standard benchmark data. Starting from existing pre-trained mesh fitting methods, we then generate inference-time mesh fits with our generalization that result in substantial improvements over the baseline pre-trained models. Finally, since we are able to generate mesh parameters for datasets that have only 2D annotations, we can retrain CNN models previously trained using only 2D ground truth. We show this retraining results in substantial performance improvements over the corresponding baseline models. Given our method is agnostic to the kind of pre-trained model used, in all experiments, we show results with various kinds of contemporary state-of-the-art methods, demonstrating its flexibility.

\begin{figure}[h!]
\centering
\includegraphics[width=1.0\linewidth]{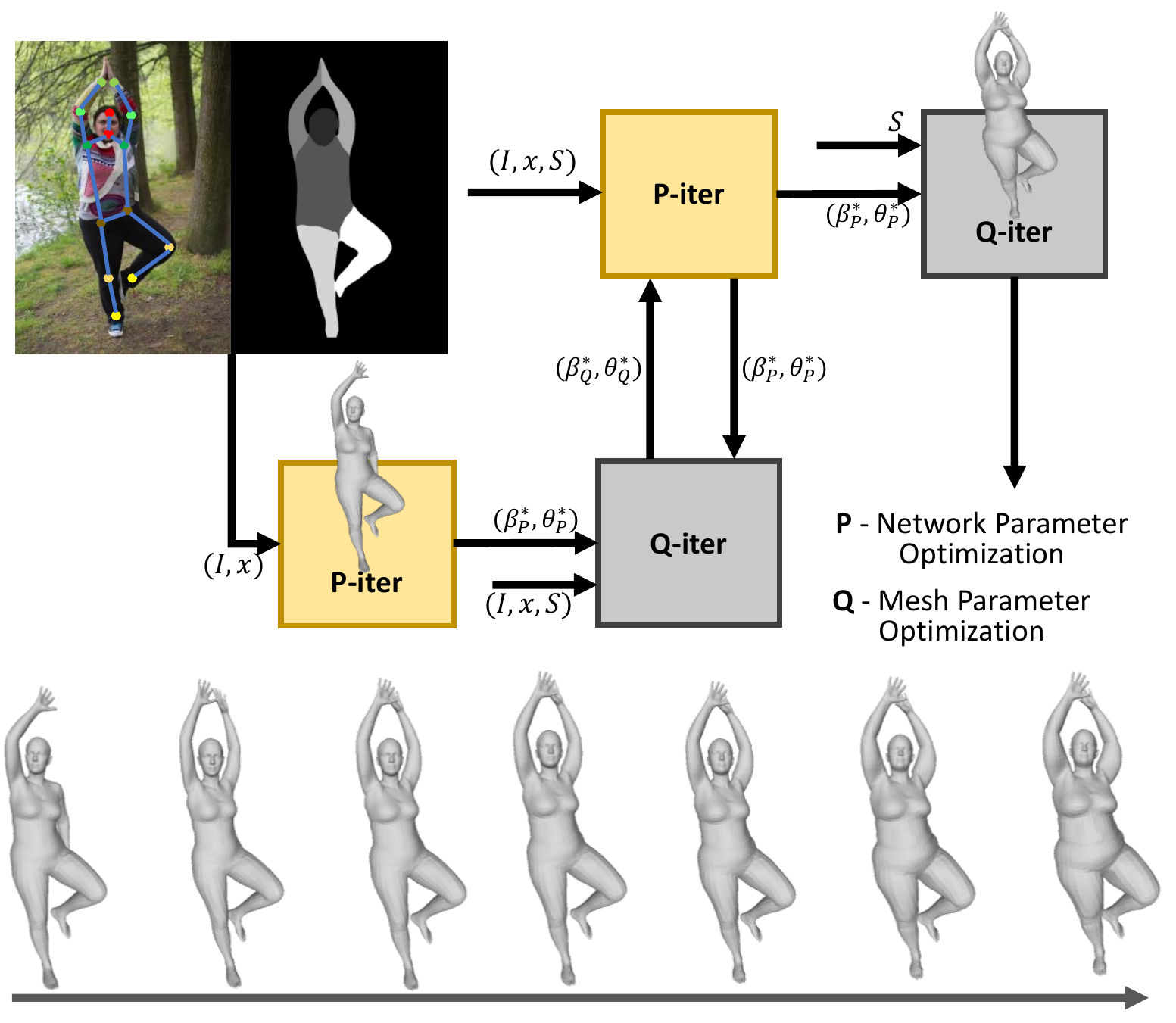}
\caption{Our proposed OMR framework. Given an image and its 2D annotations, we jointly optimize both CNN parameters and SMPL parameters iteratively, with per-step incremental updates leading to the final refined mesh.} 
\label{fig:pipeline}
\end{figure}

\section{Related Work}
\label{sec:related}
There is much recent work in human pose estimation, including estimating 2D keypoints \cite{newell2016stacked, cao2019openpose, xiao2018simple, sun2019deep, jin2020differentiable, zhang2020distribution}, 3D keypoints \cite{li20143d, tome2017lifting, pavlakos2017coarse, pavlakos2018ordinal, martinez2017simple, habibie2019wild, zhou2019hemlets, iqbal2020weakly, zeng2020srnet}, and a full human mesh \cite{bogo2016keep, kanazawa2018end, pavlakos2018learning, kolotouros2019convolutional, kolotouros2019learning, xiang2019monocular, pavlakos2019texturepose, georgakis2020hierarchical, moon2020i2l, jiang2020coherent, sengupta2020synthetic, zanfir2020weakly, kundu2020appearance, rueegg2020chained, zhang2020perceiving}. Here, we discuss methods that are directly relevant to our specific problem- fitting a 3D mesh to an input image.  \\
\indent \textbf{Direct human mesh estimation.} 
With the increasing adoption of parametric human mesh models such as the SMPL \cite{loper2015smpl}, much recent focus has been on the direct regression of the SMPL parameters that best explain the pose and shape of the person in the input image. Following the end-to-end regression design of  Kanazawa \etal \cite{kanazawa2018end}, much effort has been expended in novel architectures, with graph-based \cite{kolotouros2019convolutional}, structure-based \cite{georgakis2020hierarchical}, and even video-based \cite{arnab2019exploiting,kocabas2020vibe} approaches being some notable examples. However, as noted in Section~\ref{sec:intro}, these and other related \cite{kolotouros2019learning} methods produce biased (see Section~\ref{sec:sotaIssues} for more) and unsatisfactory results on obese person data while also requiring 3D mesh (pose and shape) as well as 2D keypoint annotations during training.  In contrast, our formulation addresses this issue of bias with a generic optimization framework and requires only 2D keypoints, which are inexpensively obtainable. Starting from such 2D points, our method learns an optimization routine to obtain the 3D pose and shape parameters that best explain the input data. Since these mesh parameters can be then used to train any of the above techniques, our method can be considered to be a flexible drop-in.

\indent \textbf{3D lifting from 2D keypoints.}
There has been some prior work in ``lifting" 2D keypoints to 3D data, with approaches based on a direct learning of 2D-3D mapping \cite{martinez2017simple, moreno20173d, biswas2019lifting, kangexplicit} and a nearest neighbor search in a database of 2D projections \cite{chen20173d, iqbal2018dual, pons2014posebits} being representative examples. However, a key difference is our method is able to exploit the context provided by a full image (as opposed to only 2D keypoints) and recover a complete body mesh.

An early method to fit the full body mesh given 2D keypoints was presented in Bogo \etal \cite{bogo2016keep} where a cost function based on the 2D reprojection loss and pose/shape priors was optimized. Kolotouros \etal \cite{kolotouros2019learning} took a different approach, using a pre-trained mesh regressor as an initializer to generate pseudo annotations with the method of Bogo \etal \cite{bogo2016keep}, which were then used to finetune the mesh regressor. Choi \etal \cite{choi2020pose2mesh} employed a two-step approach, first lifting 2D keypoints to 3D and then using the 3D keypoints to fit a full 3D mesh. Despite promising results, these fitting methods essentially rely only on 2D keypoints without using the larger context information provided by the full input image. This issue was partially addressed by Joo \etal \cite{joo2020exemplar}, where the parameters of a pre-trained CNN were optimized at test time given 2D keypoints. The optimized CNN was then used to regress the mesh parameters. While these methods present alternative views (one optimizes for mesh and the other optimizes CNN parameters), we take a more holistic view, arguing that optimizing for both parameter sets jointly leads to substantially improved fits.  

\section{Method}

\subsection{Parametric Human Body Representation}
\label{sec:smpl}
We use the Skinned Multi-Person Linear (SMPL) model \cite{loper2015smpl} to represent the 3D human mesh. SMPL factorizes the mesh parameters into a shape vector $\bm{\beta} \in \mathbb{R}^{10}$ (first 10 coefficients of the PCA of a shape space) and a pose vector $\bm{\theta} \in \mathbb{R}^{72}$ (global orientation and relative rotation angles of 23 joints in the axis-angle format).
Given these parameters, and a fixed pre-trained parameter set $\bm{\psi}$, SMPL defines a mapping $M(\bm{\beta},\bm{\theta},\bm{\psi}): \mathbb{R}^{82}\rightarrow \mathbb{R}^{3N}$ from the 82-dimensional parameter space to the space of $N=6890$ 3D mesh vertices that represent the particular pose and shape of the human body. Given these N vertices $\bm{J} \in \mathbb{R}^{3N}$, the $K=24$ joints $\bm{X} \in \mathbb{R}^{3K}$ defined by the model are obtained as $\bm{X}=\bm{W}\bm{J}$, where $\bm{W}$ is a learned joint regression matrix. These 3D joints can then be used to obtain the 2D image points $\bm{x} \in \mathbb{R}^{2K}$ with a known camera model, e.g., a weak-perspective model \cite{kanazawa2018end}: $\bm{x}=s\Pi(\bm{X}(\bm{\beta},\bm{\theta})) + \bm{t}$, where $\bm{t}\in \mathbb{R}^2$ and $s\in \mathbb{R}$ are translation and scale, and $\Pi$ is an orthographic projection. Therefore, the complete recovery of the 3D mesh corresponding to a person image involves estimating the set of parameters $\bm{\Theta}=\{\bm{\beta},\bm{\theta},s,\bm{t}\}$.

\subsection{Current Open Problems and Biased Estimators}
\label{sec:sotaIssues}
Given a training set of $n$ images $\mathbf{I}_{i}, i=1, \ldots, n$ and their associated parametric annotations $\Theta_{i}, i=1, \ldots, n$, the currently dominant paradigm for human mesh recovery is to employ a regression-based approach where the model is trained with supervision from the accompanying parametric annotations. This typically takes the form of a Euclidean loss between the ground-truth and the predicted parameter vectors \cite{kanazawa2018end,kolotouros2019convolutional,georgakis2020hierarchical}. However, as noted in Section~\ref{sec:intro}, these parametric mesh annotations are either scarcely available (e.g., very few public datasets come with these annotations) or are very expensive to create (e.g., see Loper \etal \cite{loper2014mosh}). These issues are only exacerbated for obese person images with little-to-no data available for training obese mesh estimators. On the other hand, 2D keypoint annotations can be obtained rather inexpensively (e.g., simply clicking on points on images). If we are able to generate reasonably reliable 3D mesh estimates from these 2D keypoints, we can automatically create mesh annotations readily available for retraining existing human mesh recovery models. With the current state-of-the-art methods biased due to lack of relevant annotated obese person data (see results in Figure~\ref{fig:teaser}), such retraining strategies help crucially in taking a step towards unbiased mesh estimators that work well for both obese and general images (see our results in Figure~\ref{fig:teaser}). 

An early approach to address the issue of generating mesh estimates from 2D keypoints was proposed in Bogo \etal \cite{bogo2016keep}, where a cost function comprising the 2D reprojection error and associated pose and shape priors was optimized for the mesh parameters $\bm{\Theta}$. Concretely, this optimization problem can be written as:

\begin{equation}
    \mathbf{\Theta}^{*}=\underset{\mathbf{\Theta}}{\arg\min} ~L_{2D}(\mathbf{x},\hat{\mathbf{x}})
    \label{eq:SMPLify}
\end{equation}

where $L_{2D}(\mathbf{x},\hat{\mathbf{x}})$ measures the deviation of the predicted $r$ 2D keypoints $\hat{\mathbf{x}} \in \mathbb{R}^{r \times 2}$ from its ground truth set $\mathbf{x} \in \mathbb{R}^{r \times 2}$. 

With an appropriate starting point and priors for pose $\theta$ and shape $\beta$, this optimization problem does converge, giving fairly reasonable values for $\bm{\Theta}=\{\beta,\theta,s,t\}$ \cite{bogo2016keep}. 

There are, however, a number of issues with this formulation for both obese as well as general mesh fitting. First, the prior terms used in this cost function are not representative of obese person data \cite{kanazawa2018end}, making the resulting mesh estimator biased towards only non-obese data. Given the same priors are used in many recent direct-regression-based follow-up methods \cite{kolotouros2019learning,kocabas2020vibe,georgakis2020hierarchical}, this is a crucial currently-unaddressed problem. Next, as established in recent work \cite{kolotouros2019learning}, the optimization results depend on good initialization, which is not trivial to determine particularly for obese or in-the-wild images. Finally, while minimizing the reprojection error based on 2D keypoints can ideally lead to perfect 2D fits (i.e., 2D loss is zero), the resulting $\bm{\Theta}$ can still be off due to the classic depth ambiguity problem (multiple 3D configurations being able to explain the same 2D projection).

While recent follow-up optimization approaches, e.g., EFT \cite{joo2020exemplar}, attempts to address some of these issues (e.g., depth ambiguity), the crucial problem of biased estimation remains (see EFT results in Figure~\ref{fig:teaser}). Specifically, given a pre-trained mesh regressor (e.g., HMR \cite{kanazawa2018end}) $\bm{\Phi}:\mathbb{R}^{M\times N \times 3}\to \mathbb{R}^{d}$ trained to predict $\bm{\Theta} \in \mathbb{R}^{d}$, typically realized as a CNN), this method optimizes a similar 2D reprojection objective with the difference being parameters of optimization are the CNN parameters (instead of the mesh parameters $\bm{\Theta}$ above). Representing all parameters of the CNN $\bm{\Phi}$ as the vector $\bm{\alpha}$, the optimization problem is:

\begin{equation}
    \bm{\alpha}^* = \underset{\bm{\alpha}}{\arg\min} ~L_{2D}(\pi f(\bm{\Phi}(\bm{I})), x)
    \label{eq:EFT}
\end{equation}

where the function $f$ is a composition of the functions that map (a) the mesh parameters $\bm{\Theta}$ to vertices $\mathbf{V}$ and (b) the vertices $\mathbf{V}$ to 3D joints $\mathbf{X}$ (see Section~\ref{sec:smpl}), and $\pi$ represents the camera model that is used to project to 2D points. Given $\bm{\alpha}^*$, and hence the CNN $\bm{\Phi}^*$, the mesh parameters for the image $\bm{I}$ are then obtained as $\bm{\Theta}^*=\bm{\Phi}^*(\bm{I})$.

Since this formulation relies on a pre-trained mesh estimator (e.g., HMR \cite{kanazawa2018end}), the problem of depth ambiguity can be alleviated to a certain extent with such a data-driven model, albeit restricted to the domain of the training data. Crucially, however, these pre-trained models also rely on the same priors as above, resulting in the same issue of biased estimation for obese data as above. Bias is not the only problem with EFT. Given that one needs a large number of iterations to obtain good fits, and that the objective comprises only a 2D term, this leads to the risk of overfitting the 2D cost function, which is both generally undesirable and particularly detrimental to obese mesh fitting where the 2D loss can be zero while resulting in an incorrect shape. 

\subsection{Proposed Method}
We take a structured approach to address the issues discussed above. We first propose learnable 2D shape constraints that can be easily integrated into the objective functions of the approaches discussed above, leading immediately to unbiased mesh estimation for obese person images. We then propose a generalized algorithm that optimizes for both mesh and CNN parameters, leading to a holistic optimization-based mesh fitting technique and improved (w.r.t. corresponding base methods) mesh fits for standard benchmark images. We show how this addresses shortcomings of each individual parameter optimization strategy above while also resulting in a flexible framework for integrating our proposed 2D shape constraints.

\subsubsection{Trainable 2D shape constraints}

As discussed above, optimizing a purely 2D keypoints-based loss function is insufficient to retrieve the correct shape since one can fit the 2D loss even while the 3D shape is incorrect. Furthermore, the priors embedded (either implicitly as in EFT \cite{joo2020exemplar} or explicitly as in Bogo \etal \cite{bogo2016keep}) are themselves biased and insufficiently representative to cover both obese and non-obese data (leading to the failure cases of Figure~\ref{fig:teaser}). To address this issue of bias, we propose a 2D shape loss term that can be easily added to the training objectives of Equation~\ref{eq:SMPLify} or~\ref{eq:EFT}. Specifically, we manually annotate images in our internally collected obese person data with part-based segmentation labels following the six-part segmentation strategy (including head, torso, left/right arms and left/right legs) in the LSP dataset  \cite{johnson2010clustered,johnson2011learning}. Given this binary mask $\mathbf{S}_{i}$ ($i=1,\cdots,6$ for the six parts) for an image $\mathbf{I}$, we define our 2D shape loss as:

\begin{equation}
    L_{\text{shape}}=\sum_{i=1}^{6}1-\frac{\sum_{m,n}\mathbf{\hat{S}}_{i}^{m,n} \cdot \mathbf{S}_{i}^{m,n}}{\sum_{m,n}\mathbf{\hat{S}}_{i}^{m,n} + \mathbf{S}_{i}^{m,n} - \mathbf{\hat{S}}_{i}^{m,n} \cdot \mathbf{S}_{i}^{m,n}}
\end{equation}
where $\mathbf{\hat{S}_{i}}^{m,n}$ is the $(m,n)$ pixel of the mask estimated during the course of optimization. Note this can be easily obtained after the mesh vertices are computed based on the estimated $\bm{\Theta}$. Given $L_{\text{shape}}$, we can easily adapt Equations~\ref{eq:SMPLify} and~\ref{eq:EFT} as:

\begin{align*} 
\mathbf{\Theta}^{*}=\underset{\mathbf{\Theta}}{\arg\min} ~L_{2D}(\mathbf{x},\hat{\mathbf{x}})+L_{\text{shape}}\\ 
\bm{\alpha}^* = \underset{\bm{\alpha}}{\arg\min} ~L_{2D}(\pi f(\bm{\Phi}(\bm{I})), x)+L_{\text{shape}}
\label{eq:optSeg}
\end{align*} 

\subsubsection{Generalizing mesh and CNN optimization}
While our shape constraints help alleviate the bias issues, they do not help tackle the aforementioned problems of depth ambiguity and overfitting. In order to alleviate these issues while also being able to generate unbiased meshes for both obese and non-obese data, we propose \textsl{optimization for mesh recovery} (\textbf{OMR}), a generalized mesh fitting algorithm that considers both mesh parameters $\bm{\Theta}$ and model parameters $\bm{\alpha}$ as an explicit part of the optimization problem. Our core argument is two-fold: (a) the issue of depth ambiguity can be alleviated by using a data-driven predictor, and (b) the issue of overfitting can be tackled by using explicit pose and shape regularizers in the cost function. This leads to our proposed formulation that employs the classical alternating directions scheme as part of a multi-step optimization strategy in solving for the mesh parameters $\bm{\Theta}$ that best explains the pose and shape of the person in the input image. Crucially, our proposed $L_{\text{shape}}$ can also be easily integrated in this pipeline, resulting in a generalized framework for unbiased mesh estimation. Specifically, our $L_{\text{shape}}$ can help adapt the mesh fits to the specific distribution of data (this can be thought of as an explicit regularizer). It is important to note that OMR can be used without $L_{\text{shape}}$ as well in which case the alternating directions optimization scheme helps produce accurate pose vectors. 

Given $\bm{\Phi}$, $\mathbf{I}$, and $\mathbf{x}$, we first optimize a 2D reprojection loss for $\bm{\alpha}$, giving an updated CNN with the parameters:
\begin{equation}
\bm{\alpha}^* = \underset{\bm{\alpha}}{\arg\min} ~L_{2D}(\pi f(\bm{\Phi}(I)),\mathbf{x})
\label{eq:proposedStep0}
\end{equation}
Given $\bm{\Phi}^*$, we compute the $\bm{\Theta}$ prediction for the image $\mathbf{I}$ as: $\bm{\Theta}_{0}^*=[\bm{\theta}^*,\bm{\beta}^*,s^*,\bm{t}^*]=\bm{\Phi}^*(I)$. This $\bm{\Theta}_{0}^*$ is then used as initialization to solve a new optimization problem with respect to the mesh parameters:
\begin{equation}
    \bm{\Theta}_{1}^* = \underset{\bm{\Theta}}{\arg\min} ~L_{2D}(\pi M(\beta,\theta), x)+L_{\theta}(\theta)+L_{\text{shape}}
    \label{eq:proposedStep1}
\end{equation}
where $\bm{M}$ is the SMPL mapping defined in Section~\ref{sec:smpl}. This optimization order provides a good starting point, helping address the initialization issue noted in Section~\ref{sec:sotaIssues}.

\begin{figure*}[h!]
\centering
\includegraphics[scale=0.22]{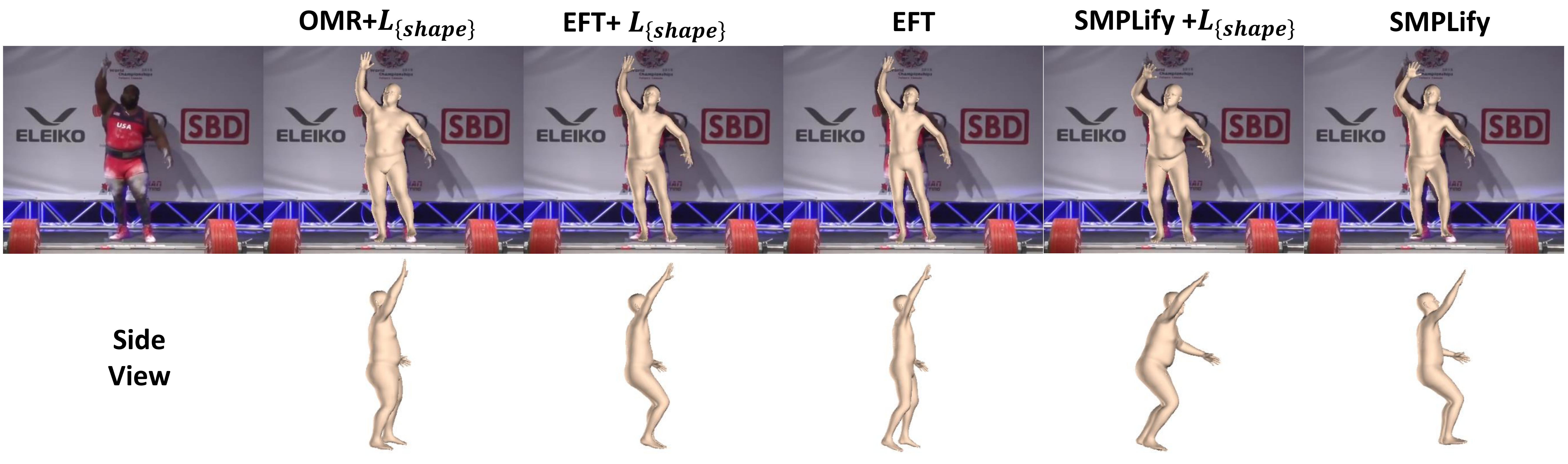}
\caption{Qualitative illustrations for improvements with our proposed $L_{\text{shape}}$.}
\label{fig:omrSmplifyEFTSegCompare}
\end{figure*}

\begin{table*}[h!]
\begin{center}
\scalebox{0.8}{
\begin{tabular}{ccc}
\toprule
SSP-3D              & PA-MPJPE (mm)     & PVE-T (mm) \\
\midrule
SPIN                    & 53.57     & 35.68  \\
\midrule
SMPLify w/o $L_\text{shape}$   & 56.78     & 31.86  \\
SMPLify+$L_\text{shape}$       & 53.23     & 28.99  \\
\midrule
EFT w/o $L_\text{shape}$        & 51.96     & 34.03  \\
EFT+$L_\text{shape}$               & 50.68     & 32.56  \\
\midrule
OMR w/o $L_\text{shape}$        & 49.67     & 32.71  \\
OMR+$L_\text{shape}$            & \textbf{46.26}     & \textbf{21.52}  \\
\bottomrule
\end{tabular}
~~
\begin{tabular}{ccccc}
\toprule
SSP-3D (PVE-T)           & Torso    & Legs    & Arms    & Head \\
\midrule
SPIN             & 53.56    & 26.15   & 38.00   & 32.16 \\
\midrule
SMPLify w/o $L_\text{shape}$    & 50.09    & 19.99   & 36.68   & 24.79 \\
SMPLify+$L_\text{shape}$    & 46.31    & 19.06   & 34.32   & 23.68 \\
\midrule
EFT w/o $L_\text{shape}$        & 54.87    & 22.54   & 38.23   & 29.11 \\
EFT+$L_\text{shape}$       & 51.99    & 21.61   & 35.08   & 29.90 \\
\midrule
OMR w/o $L_\text{shape}$            & 52.32     & 21.68    & 35.35  & 29.88\\
OMR+$L_\text{shape}$            & \textbf{26.41}     & \textbf{19.69}    & \textbf{20.70}  & \textbf{21.53}\\
\bottomrule
\end{tabular}}
\end{center}
\caption{Improving SMPLify \cite{bogo2016keep} and EFT \cite{joo2020exemplar} on SSP-3D with $L_\text{shape}$. Our OMR generalization outperforms both.}
\label{tab:sspspin}
\end{table*}

Next, we use the $\bm{\Theta}_{1}^*$ from Equation~\ref{eq:proposedStep1} as an explicit regularization to optimize the CNN parameters, which we realize by modifying the problem of Equation~\ref{eq:EFT} as:

\begin{equation}
\bm{\alpha}^* = \underset{\bm{\alpha}}{\arg\min} ~L_{2D}(\pi f(\bm{\Phi}(I)),\mathbf{x}) + \|\bm{\Theta}-\bm{\Theta}_{1}^*\|^2_2 + L_{\text{shape}}
\label{eq:proposedStep2}
\end{equation}
Given this new $\bm{\Phi}^*$, we obtain a new $\bm{\Theta}$ prediction as $\bm{\Theta}_{2}^*=[\bm{\theta}^*,\bm{\beta}^*,s^*,\bm{t}^*]=\bm{\Phi}^*(I)$. This new $\bm{\Theta}_{2}^*$ can now be used to solve a new optimization problem of Equation~\ref{eq:proposedStep1} above, whose solution can be used to solve a new optimization problem of Equation~\ref{eq:proposedStep2}, thereby leading to an iterative alternating optimization of $\bm{\Theta}$ and $\bm{\alpha}$. 

OMR addresses all limitations of prior work in a principled manner. First, the result of step 0 ensures a good initialization for step 1. Since this only depends on the pre-trained CNN $\bm{\Phi}$, OMR can be used as a drop-in to improve any pre-trained model's performance (e.g., we show results with SPIN \cite{kolotouros2019learning}, CMR \cite{kolotouros2019convolutional}, and HKMR \cite{georgakis2020hierarchical}). Second, our step 1 provides explicit regularization to address EFT's overfitting issue. Finally, OMR is flexible to be optimized with our proposed 2D shape constraints, resulting in a framework that provides accurate fits for both obese as well as general data, leading to reduced bias. Note that since OMR starts with a $P$-iteration solution for Equation~\ref{eq:proposedStep0} and subsequently alternates between an $Q$-iteration Equation~\ref{eq:proposedStep1} and an $P$-iteration Equation~\ref{eq:proposedStep2}, we use the notation $(n+1)PnQ$ to refer to the number of OMR steps.

\section{Experiments and Results}
\label{sec:exp}
In this section, we discuss the results of a number of experiments we conducted to demonstrate the efficacy of both our shape constraints as well as OMR. First, we show how our 2D shape loss improves the performance of existing methods, taking a step towards unbiased estimation for obese data. Next, we show how our generalization, i.e., OMR, gives better mesh estimates across both obese and non-obese data when compared to existing optimization strategies. Finally, we show how OMR can generate $\bm{\Theta}$ annotations for datasets that only have 2D annotations, and how they can be used to retrain and improve the performance of existing mesh recovery models.

\subsection{Datasets, Evaluation, and a New Metric}
For datasets with only 2D keypoint annotations, we use LSP \cite{johnson2010clustered}, LSP-extended \cite{johnson2011learning}, MPII \cite{andriluka20142d} and MS COCO \cite{lin2014microsoft}. For datasets with both 2D and 3D keypoint annotations, we use MPI-INF-3DHP \cite{mehta2017monocular} and Human3.6M \cite{ionescu2013human3}. Furthermore, to demonstrate results on obese person data, we use SSP-3D \cite{STRAPS2018BMVC} as well as an internally collected (by scraping the web and manually filtering) \texttt{LargeSize} dataset. While SSP-3D has a varied set of annotated images, we only use data with extreme shape parameters for obese person evaluation, whereas our LargeSize dataset has 2D keypoints and body-part segmentation masks. 

\begin{figure}[h!]
\centering
\includegraphics[scale=0.6]{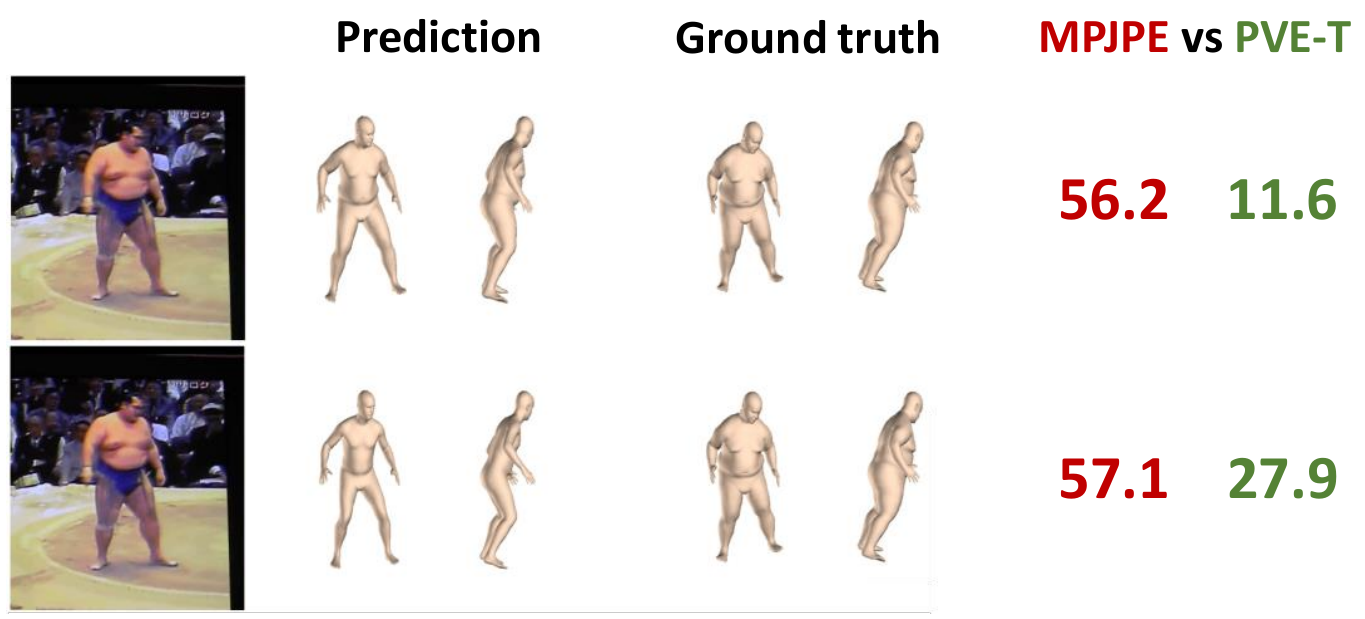}
\caption{MPJPE vs. proposed PVE-T.}
\label{fig:metricCompare}
\end{figure}

For all experiments, we report results with the standard mean per-joint-position-error (MPJPE) metric and its procrustes-aligned variant (PA-MPJPE) \cite{kanazawa2018end,kolotouros2019learning}. Since this metric only measures deviation between a set of sparse keypoints, it is insufficient to quantify shape errors. To address this issue, we propose a new metric, called per-vertex-error-T-pose (PVE-T). Given two shape vectors, PVE-T first computes one mesh corresponding to each vector (by setting the pose to mean pose, i.e., zero vector). Given the two meshes, it considers all pairs of corresponding vertices, measures their deviation using the Euclidean metric, and returns a mean over all these values. As can be seen from Fig~\ref{fig:metricCompare}, PVE-T helps capture shape errors more representatively when compared to MPJPE (e.g., the MPJPE values are almost similar, whereas the PVE-T values are more different, thereby being more representative of the deviations between the prediction and the ground truth). 

\subsection{Evaluating Shape Constraints}
We first present results with our proposed baseline shape constraints when used with SMPLify \cite{bogo2016keep} and EFT \cite{joo2020exemplar}. While we use SPIN \cite{kolotouros2019learning} as the base model, our method is applicable to and improves the performance of other methods as well (please see supplemental material for these results). Table~\ref{tab:sspspin} shows results on the SSP-3D dataset where we see the proposed $L_{\text{shape}}$ loss consistently reduces both PA-MPJPE and PVE-T errors across all the methods (the right part of the table shows per-body-part PVE-T values to help understand local shape improvements with $L_{\text{shape}}$). Crucially, $L_{\text{shape}}$ helps even a relatively weaker baseline (SMPLify) outperform SPIN, a popular state-of-the-art method, on the PA-MPJPE metric.  Furthermore, the proposed OMR generalization outperforms both SMPLify and EFT with and without $L_{\text{shape}}$ while also substantially reducing the error w.r.t. SPIN (46.26 mm PA-MPJPE vs. 53.57 for SPIN). 

Finally, we show some qualitative results in Figure~\ref{fig:omrSmplifyEFTSegCompare} where one can note how $L_{\text{shape}}$ helps improve the shape fits of EFT and SMPLify (both of which generate good pose fits but not shape, which is addressed by $L_{\text{shape}}$). Crucially, while $L_{\text{shape}}$ helps improve EFT's shape, this results in the pose getting degraded (see side view, columns 3 and 4); this is the case with SMPLify as well (last two columns). This problem is clearly addressed by our OMR generalization where we see much better fits for both pose and shape. This is a direct consequence of our alternating directions optimization design, where the $P$-iteration explores reasonable pose configurations and provides better initialization parameters for the $Q$-iteration step, which in turn recovers better shape explanations to guide the $P$-iteration step.

\begin{table}[h!]
\begin{center}
\scalebox{0.65}{
\begin{tabular}{ccc}
\toprule
Human3.6M              & MPJPE     & PA-MPJPE \\
\midrule
SPIN \cite{kolotouros2019learning}                  & 64.95     & 43.78  \\
\midrule
SMPLify - 20           & 71.99     & 45.17  \\
SMPLify - 100          & 82.90     & 50.23  \\
\midrule
EFT - 20               & 61.24     & 40.82  \\
EFT - 100              & 63.26     & 37.95  \\
\midrule
OMR (1P1Q)            & 63.18     & 40.80  \\
OMR (5P4Q)            & \textbf{61.07}     & \textbf{37.70}  \\
\bottomrule
\end{tabular}
~~
\begin{tabular}{ccc}
\toprule
Human3.6M              & MPJPE     & PA-MPJPE \\
\midrule
HKMR \cite{georgakis2020hierarchical}                  & 65.03     & 46.53  \\
\midrule
SMPLify - 20           & 69.57     & 44.61  \\
SMPLify - 100          & 81.66     & 50.18  \\
\midrule
EFT - 20               & 58.41     & 39.53  \\
EFT - 100              & 66.62     & 41.81  \\
\midrule
OMR (1P1Q)            & 59.15     & 39.25  \\
OMR (5P4Q)            & \textbf{56.97}     & \textbf{38.60}  \\
\bottomrule
\end{tabular}}
\end{center}
\caption{SMPLify vs. EFT vs. OMR on Human3.6M using SPIN and HKMR base models. All numbers in mm.}
\label{tab:acrossModel}
\end{table}

\begin{table*}[h!]
  \begin{center}
  \scalebox{0.7}{
    \begin{tabular}{ccccc}
    \toprule
     SSP-3D    &   MPJPE  &  PA-MPJPE  &    PVE-T  &  mIoU  \\
    \midrule
     SPIN \cite{kolotouros2019learning}  & 92.03 & 53.57  & 35.68 & 0.6570 \\
     SPIN - SMPLify                      & 106.50 & 59.45  & 32.65 & 0.6889 \\
     SPIN - EFT                          & 88.60 & 55.14  & 33.05 & 0.6823 \\
     SPIN - OMR                          & \textbf{84.48} & \textbf{50.16} & \textbf{27.36} & \textbf{0.7088} \\
     \midrule
     HKMR \cite{georgakis2020hierarchical}  & 98.02 & 57.53  & 34.54 & 0.6647 \\
     HKMR - SMPLify                         & 107.55 & 60.63  & 32.62 & 0.6799 \\
     HKMR - EFT                             & 92.53 & 56.31  & 32.59 & 0.6825 \\
     HKMR - OMR                             & \textbf{88.81} & \textbf{52.43}  & \textbf{27.23} & \textbf{0.7028} \\
    \bottomrule
    \end{tabular}}
    ~~
    \scalebox{0.7}{
    \begin{tabular}{ccccc}
    \toprule
      Human3.6M   & \multicolumn{2}{c}{Protocol \#1}  &  \multicolumn{2}{c}{Protocol \#2}  \\
    \cline{2-5}
         &   MPJPE  &  PA-MPJPE  &    MPJPE  &  PA-MPJPE  \\
    \midrule
     SPIN \cite{kolotouros2019learning}                 & 65.60 & 44.1  & 62.23 & 41.1 \\
     SPIN - SMPLify                                     & 65.12 & 45.2  & 61.39 & 42.6 \\
     SPIN - EFT                                         & 63.40 & 44.3  & 60.00 & 41.5 \\
     SPIN - OMR       & \textbf{61.95} & \textbf{43.7} & \textbf{58.51} & \textbf{41.0} \\
     \midrule
     HKMR \cite{georgakis2020hierarchical}              & 64.02 & 45.9  & 59.62 & 43.2 \\
     HKMR - SMPLify                                     & 65.91 & 47.3  & 62.22 & 44.4 \\
     HKMR - EFT                                         & 64.04 & 46.3  & 62.22 & 43.4 \\
     HKMR - OMR    & \textbf{62.70} & \textbf{45.6}  & \textbf{59.36} & \textbf{42.9} \\
    \bottomrule
    \end{tabular}}
  \vspace{-2.5mm}
  \end{center}
  \caption{Improving baseline models by retraining with annotations generated by our method.}
      \label{tab:Retraining}
\end{table*}

\begin{table*}[h!]
\begin{center}
\scalebox{0.7}{
    \begin{tabular}{ccc}
    \toprule
     \multirow{2}{3cm}{\centering \textbf{Human3.6M}}    &  \multicolumn{2}{c}{Protocol \#2}  \\
    \cline{2-3}
         &   MPJPE  &  PA-MPJPE  \\
    \midrule
     HMR \cite{kanazawa2018end}                  & 88.0  & 56.8 \\
     CMR \cite{kolotouros2019convolutional}      & 71.9  & 50.1 \\
     SPIN \cite{kolotouros2019learning}          & 62.23 & 41.1 \\
     HKMR \cite{georgakis2020hierarchical}       & 59.62 & 43.2 \\
     Pose2Mesh \cite{choi2020pose2mesh}          &  64.9  & 47.0 \\
     \midrule
     HKMR - OMR    & 59.36 & 42.9 \\
     SPIN - OMR       & \textbf{58.51} & \textbf{41.0} \\
    \bottomrule
    \end{tabular}
    ~~
    \scalebox{1.1}{
    \begin{tabular}{cc}
    \toprule
    \textbf{MPI-INF-3DHP}     &   MPJPE \\
    \midrule
     Mehta \etal\cite{mehta2017monocular}                             & 117.6 \\
     VNect \cite{mehta2017vnect}                             & 124.7   \\
     HMR \cite{kanazawa2018end}              & 124.2 \\
     HKMR \cite{georgakis2020hierarchical}   & 108.9   \\
     SPIN \cite{kolotouros2019learning}      & 105.2  \\
     \midrule
     HKMR - OMR    & \textbf{100.1}  \\
     SPIN - OMR    & 100.9 \\
    \bottomrule
    \end{tabular}}
~~
    \begin{tabular}{cc}
    \toprule
    \textbf{3DPW}     &   PA-MPJPE \\
    \midrule
     HMR \cite{kanazawa2018end}              & 81.3 \\
     CMR \cite{kolotouros2019convolutional}  & 70.2 \\
     HKMR \cite{georgakis2020hierarchical}   & 76.7   \\
     SPIN \cite{kolotouros2019learning}      & 59.2  \\
     Pose2Mesh \cite{choi2020pose2mesh}      & 58.9 \\
     I2LMeshNet \cite{moon2020i2l}           & 57.7 \\
     \midrule
     HKMR - OMR    & \textbf{56.1}  \\
     SPIN - OMR    & 56.5 \\
    \bottomrule
    \end{tabular}
~~
    \scalebox{1.1}{
     \begin{tabular}{ccc}
    \toprule
\textbf{LSP}     &   FB acc. & Part acc. \\
    \midrule
     Oracle \cite{bogo2016keep}         & 92.17& 88.82 \\
     SMPLify \cite{bogo2016keep}                & 91.89  & 87.71  \\
     HMR \cite{kanazawa2018end}                 & 91.67 & 87.12 \\
     SPIN \cite{kolotouros2019learning}         & 91.83 & 89.41  \\
     HKMR \cite{georgakis2020hierarchical}      & 92.23 &  89.59  \\
     \midrule
     HKMR - OMR    & \textbf{92.44} & \textbf{89.86}  \\
     SPIN - OMR & 92.35 & 89.76  \\
    \bottomrule
    \end{tabular}}}
\vspace{-2mm}
\end{center}
\caption{Comparison with competing state-of-the art methods on Human3.6M, MPI-INF-3DHP, and LSP.}
\label{tab:sota}
\end{table*}

\subsection{Generalized Model Fitting Evaluation} \label{sec:acrossDataset}
We next evaluate the generalizability of our proposed OMR in its ability to be used in conjunction with multiple existing human mesh methods. To this end, we start with existing pre-trained base models (CMR \cite{kolotouros2019convolutional}, SPIN \cite{kolotouros2019learning}, and HKMR \cite{georgakis2020hierarchical} in Table~\ref{tab:acrossModel}) (CMR results are in supplementary material) and run the three kinds of optimization algorithms (SMPLify, EFT, and OMR) on test images from the Human3.6M dataset to obtain the $\bm{\Theta}$ parameters, and consequently 3D mesh keypoints (see Section~\ref{sec:smpl}).  Note that while 3D keypoints ground truth are available, we do not use them in any capacity during the optimization process (i.e., they are only used for reporting evaluation metrics). We repeat this for all images in the evaluation set and report average error values. As can be noted from Table~\ref{tab:acrossModel}, for Human3.6M, increasing the number of iterations (from 20 to 100) in both SMPLify and EFT leads to overfitting (note increasing MPJPE and reconstruction errors), whereas OMR is able to address this issue with a steady decrease in both MPJPE and reconstruction errors. 

Note that OMR's 5P4Q strategy gives the lowest errors that are each substantially better than the corresponding base model's performance (e.g., in column 2, 64.95 mm for SPIN vs. 61.07 mm for OMR; notice similar trends for CMR and HKMR), suggesting OMR's flexibility to be used as a drop-in optimization strategy across multiple different mesh estimation techniques (see supplementary material for more results). Finally, the strong performance of OMR (w.r.t. SMPLify/EFT for various base methods) across both Tables~\ref{tab:sspspin} and~\ref{tab:acrossModel} suggest its generalizability for both (specific) obese mesh fitting and (generic) non-obese mesh fitting. See supplementary material for some qualitative results across both these cases.

\begin{figure}[h!]
\centering
\includegraphics[scale=0.365]{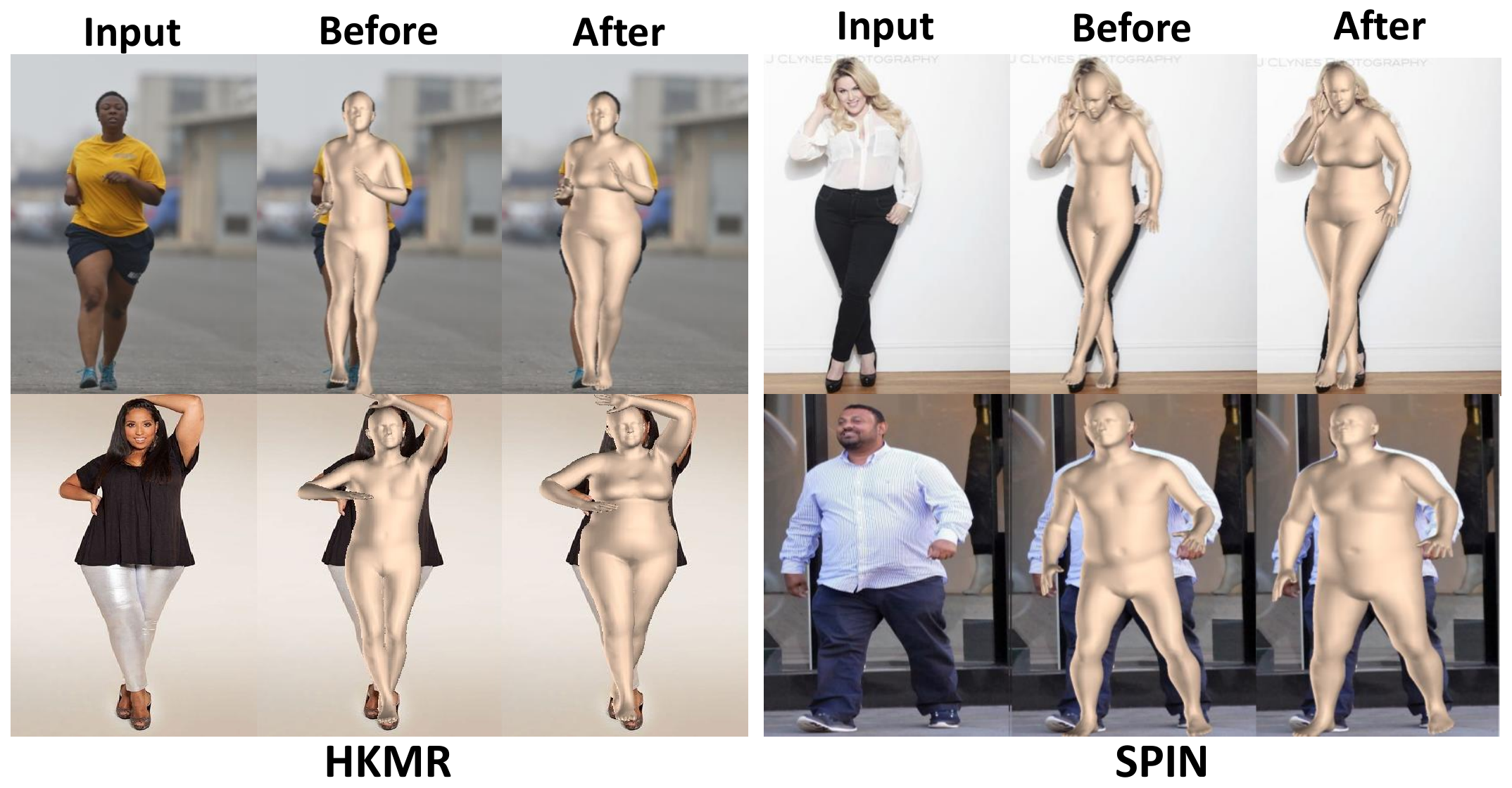}
\caption{HKMR and SPIN before and after retraining.}
\label{fig:hkmrSpinObeseRetrain}
\end{figure}

\subsection{Generating Annotations and Model Retraining} 
\label{sec:retrain}
As noted previously, current state-of-the-art methods fail to fit accurate meshes to obese data, leading to biased estimation (see Section~\ref{sec:sotaIssues} and Figure~\ref{fig:teaser}). Furthermore, as noted in Section~\ref{sec:intro}, while accurate mesh fitting needs 3D parameter supervision, it is expensive to generate these. To address both these issues, we use our proposed OMR algorithm and $L_{\text{shape}}$ to automatically generate $\bm{\Theta}$ parameters for LargeSize, LSP, LSP-extended, MPII and MSCOCO. We then retrain state-of-the-art methods with these automatically generated parameters. To demonstrate how this leads to reduced bias on obese data and improved standard benchmark performance, we use  HMR \cite{kanazawa2018end}, CMR \cite{kolotouros2019convolutional}, SPIN \cite{kolotouros2019learning} and HKMR \cite{georgakis2020hierarchical}. The corresponding default training configurations are used for a fair evaluation. Figure~\ref{fig:hkmrSpinObeseRetrain} shows substantially improved shape fits with HKMR and SPIN when compared to their corresponding baseline versions, qualitatively demonstrating the impact of our proposed method. To quantify these gains, in Table~\ref{tab:Retraining}, we show results on the standard benchmark Human3.6M dataset and the obese SSP-3D dataset, where one can note the substantial performance improvements after retraining across both obese and standard data as well as all the baseline methods (see supplementary for results with HMR and CMR). Furthermore, retraining with OMR-generated parameters leads to much better performance when compared to the corresponding SMPLify and EFT versions (see SPIN and HKMR in Table~\ref{tab:Retraining}), further validating OMR's optimization design. 

\textbf{Comparison with the state of the art.} We finally compare the performance of models retrained OMR-generated annotations with the existing state of the art. Table~\ref{tab:sota} shows our results on Human3.6M, MPI-INF-3DHP, 3DPW and LSP test sets, where one can note clear performance improvements. For instance, SPIN-OMR obtains the lowest error on Human 3.6M whereas HKMR-OMR obtains the highest foreground and part segmentation accuracy on LSP.

\section{Summary}
In this work, we considered the problem of human mesh recovery with a particular emphasis on mesh estimation for images of obese people. We noted that the current state-of-the-art methods produce biased estimates for obese images, discussed our reasoning behind this issue, and proposed ways to overcome this problem. Specifically, we first proposed new 2D shape constraints that can be flexibly used in conjunction with existing mesh fitting algorithms. We showed how this results in an immediate improvement in baseline performance. We then proposed a generalized mesh fitting algorithm, called OMR, that optimizes a reprojection error cost function in a space of both mesh parameters and CNN model parameters, showing how this results in a holistic approach that addresses the limitations of existing mesh optimization algorithms. We then showed the proposed 2D shape constraints can be easily integrated into OMR while also having the flexibility to be used with any contemporary regression-based mesh recovery algorithm. We demonstrated the efficacy of our algorithms by means of extensive experiments on both obese person data and standard benchmark data, establishing new baseline results for obese mesh recovery and state-of-the-art performance of benchmark human mesh recovery.

{\small
\bibliographystyle{ieee_fullname}
\bibliography{egbib}

\begin{thebibliography}{10}

\bibitem{karanam2020towards}
Srikrishna Karanam, Ren Li, Fan Yang, Wei Hu, Terrence Chen, and Ziyan Wu.
\newblock Towards contactless patient positioning.
\newblock {\em IEEE Transactions on Medical Imaging}, 2020.

\bibitem{ching2014patient}
William Ching, John Robinson, and Mark McEntee.
\newblock Patient-based radiographic exposure factor selection: a systematic
  review.
\newblock {\em Journal of medical radiation sciences}, 61(3):176--190, 2014.

\bibitem{kanazawa2018end}
Angjoo Kanazawa, Michael~J Black, David~W Jacobs, and Jitendra Malik.
\newblock End-to-end recovery of human shape and pose.
\newblock In {\em CVPR}, 2018.

\bibitem{kolotouros2019learning}
Nikos Kolotouros, Georgios Pavlakos, Michael~J Black, and Kostas Daniilidis.
\newblock Learning to reconstruct 3d human pose and shape via model-fitting in
  the loop.
\newblock In {\em ICCV}, 2019.

\bibitem{georgakis2020hierarchical}
Georgios Georgakis, Ren Li, Srikrishna Karanam, Terrrence Chen, Jana Kosecka,
  and Ziyan Wu.
\newblock Hierarchical kinematic human mesh recovery.
\newblock In {\em ECCV}, 2020.

\bibitem{fang2020sensitivity}
Yicheng Fang, Huangqi Zhang, Jicheng Xie, Minjie Lin, Lingjun Ying, Peipei
  Pang, and Wenbin Ji.
\newblock Sensitivity of chest ct for covid-19: comparison to rt-pcr.
\newblock {\em Radiology}, 296(2):E115--E117, 2020.

\bibitem{kass2020obesity}
David~A Kass, Priya Duggal, and Oscar Cingolani.
\newblock Obesity could shift severe covid-19 disease to younger ages.
\newblock {\em Lancet (London, England)}, 2020.

\bibitem{uMROmega}
{United Imaging uMR Omega}.
\newblock \url{https://www.dotmed.com/news/story/49443}.
\newblock Accessed: 2021-02-14.

\bibitem{loper2015smpl}
Matthew Loper, Naureen Mahmood, Javier Romero, Gerard Pons-Moll, and Michael~J
  Black.
\newblock Smpl: A skinned multi-person linear model.
\newblock {\em ACM Transactions on Graphics}, 34(6):1--16, 2015.

\bibitem{loper2014mosh}
Matthew Loper, Naureen Mahmood, and Michael~J Black.
\newblock Mosh: Motion and shape capture from sparse markers.
\newblock {\em ACM Transactions on Graphics}, 33(6):1--13, 2014.

\bibitem{johnson2010clustered}
Sam Johnson and Mark Everingham.
\newblock Clustered pose and nonlinear appearance models for human pose
  estimation.
\newblock In {\em BMVC}, 2010.

\bibitem{johnson2011learning}
Sam Johnson and Mark Everingham.
\newblock Learning effective human pose estimation from inaccurate annotation.
\newblock In {\em CVPR}, 2011.

\bibitem{andriluka20142d}
Mykhaylo Andriluka, Leonid Pishchulin, Peter Gehler, and Bernt Schiele.
\newblock 2d human pose estimation: New benchmark and state of the art
  analysis.
\newblock In {\em CVPR}, 2014.

\bibitem{lin2014microsoft}
Tsung-Yi Lin, Michael Maire, Serge Belongie, James Hays, Pietro Perona, Deva
  Ramanan, Piotr Doll{\'a}r, and C~Lawrence Zitnick.
\newblock Microsoft {COCO}: Common objects in context.
\newblock In {\em ECCV}, 2014.

\bibitem{ionescu2013human3}
Catalin Ionescu, Dragos Papava, Vlad Olaru, and Cristian Sminchisescu.
\newblock {Human3.6M}: Large scale datasets and predictive methods for 3d human
  sensing in natural environments.
\newblock {\em IEEE Transactions on Pattern Analysis and Machine Intelligence},
  36(7):1325--1339, 2013.

\bibitem{bogo2016keep}
Federica Bogo, Angjoo Kanazawa, Christoph Lassner, Peter Gehler, Javier Romero,
  and Michael~J Black.
\newblock Keep it smpl: Automatic estimation of 3d human pose and shape from a
  single image.
\newblock In {\em ECCV}, 2016.

\bibitem{joo2020exemplar}
Hanbyul Joo, Natalia Neverova, and Andrea Vedaldi.
\newblock Exemplar fine-tuning for 3d human pose fitting towards in-the-wild 3d
  human pose estimation.
\newblock {\em arXiv preprint arXiv:2004.03686}, 2020.

\bibitem{newell2016stacked}
Alejandro Newell, Kaiyu Yang, and Jia Deng.
\newblock Stacked hourglass networks for human pose estimation.
\newblock In {\em ECCV}, 2016.

\bibitem{cao2019openpose}
Z~Cao, T~Simon, SE~Wei, YA~Sheikh, et~al.
\newblock Openpose: Realtime multi-person 2d pose estimation using part
  affinity fields.
\newblock {\em IEEE Transactions on Pattern Analysis and Machine Intelligence},
  2019.

\bibitem{xiao2018simple}
Bin Xiao, Haiping Wu, and Yichen Wei.
\newblock Simple baselines for human pose estimation and tracking.
\newblock In {\em ECCV}, 2018.

\bibitem{sun2019deep}
Ke~Sun, Bin Xiao, Dong Liu, and Jingdong Wang.
\newblock Deep high-resolution representation learning for human pose
  estimation.
\newblock In {\em CVPR}, 2019.

\bibitem{jin2020differentiable}
Sheng Jin, Wentao Liu, Enze Xie, Wenhai Wang, Chen Qian, Wanli Ouyang, and Ping
  Luo.
\newblock Differentiable hierarchical graph grouping for multi-person pose
  estimation.
\newblock {\em arXiv preprint arXiv:2007.11864}, 2020.

\bibitem{zhang2020distribution}
Feng Zhang, Xiatian Zhu, Hanbin Dai, Mao Ye, and Ce~Zhu.
\newblock Distribution-aware coordinate representation for human pose
  estimation.
\newblock In {\em CVPR}, 2020.

\bibitem{li20143d}
Sijin Li and Antoni~B Chan.
\newblock 3d human pose estimation from monocular images with deep
  convolutional neural network.
\newblock In {\em ACCV}, 2014.

\bibitem{tome2017lifting}
Denis Tome, Chris Russell, and Lourdes Agapito.
\newblock Lifting from the deep: Convolutional 3d pose estimation from a single
  image.
\newblock In {\em CVPR}, 2017.

\bibitem{pavlakos2017coarse}
Georgios Pavlakos, Xiaowei Zhou, Konstantinos~G Derpanis, and Kostas
  Daniilidis.
\newblock Coarse-to-fine volumetric prediction for single-image 3d human pose.
\newblock In {\em CVPR}, 2017.

\bibitem{pavlakos2018ordinal}
Georgios Pavlakos, Xiaowei Zhou, and Kostas Daniilidis.
\newblock Ordinal depth supervision for 3d human pose estimation.
\newblock In {\em CVPR}, 2018.

\bibitem{martinez2017simple}
Julieta Martinez, Rayat Hossain, Javier Romero, and James~J Little.
\newblock A simple yet effective baseline for 3d human pose estimation.
\newblock In {\em ICCV}, 2017.

\bibitem{habibie2019wild}
Ikhsanul Habibie, Weipeng Xu, Dushyant Mehta, Gerard Pons-Moll, and Christian
  Theobalt.
\newblock In the wild human pose estimation using explicit 2d features and
  intermediate 3d representations.
\newblock In {\em CVPR}, 2019.

\bibitem{zhou2019hemlets}
Kun Zhou, Xiaoguang Han, Nianjuan Jiang, Kui Jia, and Jiangbo Lu.
\newblock {HEMlets} pose: Learning part-centric heatmap triplets for accurate
  3d human pose estimation.
\newblock In {\em ICCV}, 2019.

\bibitem{iqbal2020weakly}
Umar Iqbal, Pavlo Molchanov, and Jan Kautz.
\newblock Weakly-supervised 3d human pose learning via multi-view images in the
  wild.
\newblock In {\em CVPR}, 2020.

\bibitem{zeng2020srnet}
Ailing Zeng, Xiao Sun, Fuyang Huang, Minhao Liu, Qiang Xu, and Stephen Lin.
\newblock {SRNet}: Improving generalization in 3d human pose estimation with a
  split-and-recombine approach.
\newblock {\em arXiv preprint arXiv:2007.09389}, 2020.

\bibitem{pavlakos2018learning}
Georgios Pavlakos, Luyang Zhu, Xiaowei Zhou, and Kostas Daniilidis.
\newblock Learning to estimate 3d human pose and shape from a single color
  image.
\newblock In {\em CVPR}, 2018.

\bibitem{kolotouros2019convolutional}
Nikos Kolotouros, Georgios Pavlakos, and Kostas Daniilidis.
\newblock Convolutional mesh regression for single-image human shape
  reconstruction.
\newblock In {\em CVPR}, 2019.

\bibitem{xiang2019monocular}
Donglai Xiang, Hanbyul Joo, and Yaser Sheikh.
\newblock Monocular total capture: Posing face, body, and hands in the wild.
\newblock In {\em CVPR}, 2019.

\bibitem{pavlakos2019texturepose}
Georgios Pavlakos, Nikos Kolotouros, and Kostas Daniilidis.
\newblock {TexturePose}: Supervising human mesh estimation with texture
  consistency.
\newblock In {\em ICCV}, 2019.

\bibitem{moon2020i2l}
Gyeongsik Moon and Kyoung~Mu Lee.
\newblock I2l-meshnet: Image-to-lixel prediction network for accurate 3d human
  pose and mesh estimation from a single rgb image.
\newblock {\em arXiv preprint arXiv:2008.03713}, 2020.

\bibitem{jiang2020coherent}
Wen Jiang, Nikos Kolotouros, Georgios Pavlakos, Xiaowei Zhou, and Kostas
  Daniilidis.
\newblock Coherent reconstruction of multiple humans from a single image.
\newblock In {\em CVPR}, 2020.

\bibitem{sengupta2020synthetic}
Akash Sengupta, Ignas Budvytis, and Roberto Cipolla.
\newblock Synthetic training for accurate 3d human pose and shape estimation in
  the wild.
\newblock {\em arXiv preprint arXiv:2009.10013}, 2020.

\bibitem{zanfir2020weakly}
Andrei Zanfir, Eduard~Gabriel Bazavan, Hongyi Xu, Bill Freeman, Rahul
  Sukthankar, and Cristian Sminchisescu.
\newblock Weakly supervised 3d human pose and shape reconstruction with
  normalizing flows.
\newblock {\em arXiv preprint arXiv:2003.10350}, 2020.

\bibitem{kundu2020appearance}
Jogendra~Nath Kundu, Mugalodi Rakesh, Varun Jampani, Rahul~Mysore Venkatesh,
  and R~Venkatesh Babu.
\newblock Appearance consensus driven self-supervised human mesh recovery.
\newblock {\em arXiv preprint arXiv:2008.01341}, 2020.

\bibitem{rueegg2020chained}
Nadine Rueegg, Christoph Lassner, Michael~J Black, and Konrad Schindler.
\newblock Chained representation cycling: Learning to estimate 3d human pose
  and shape by cycling between representations.
\newblock {\em arXiv preprint arXiv:2001.01613}, 2020.

\bibitem{zhang2020perceiving}
Jason~Y Zhang, Sam Pepose, Hanbyul Joo, Deva Ramanan, Jitendra Malik, and
  Angjoo Kanazawa.
\newblock Perceiving 3d human-object spatial arrangements from a single image
  in the wild.
\newblock In {\em ECCV}, 2020.

\bibitem{arnab2019exploiting}
Anurag Arnab, Carl Doersch, and Andrew Zisserman.
\newblock Exploiting temporal context for 3d human pose estimation in the wild.
\newblock In {\em CVPR}, 2019.

\bibitem{kocabas2020vibe}
Muhammed Kocabas, Nikos Athanasiou, and Michael~J Black.
\newblock Vibe: Video inference for human body pose and shape estimation.
\newblock In {\em CVPR}, 2020.

\bibitem{moreno20173d}
Francesc Moreno-Noguer.
\newblock 3d human pose estimation from a single image via distance matrix
  regression.
\newblock In {\em CVPR}, 2017.

\bibitem{biswas2019lifting}
Sandika Biswas, Sanjana Sinha, Kavya Gupta, and Brojeshwar Bhowmick.
\newblock Lifting 2d human pose to 3d: A weakly supervised approach.
\newblock In {\em IJCNN}, 2019.

\bibitem{kangexplicit}
Yangyuxuan Kang, Anbang Yao, Shandong Wang, Ming Lu, Yurong Chen, and Enhua Wu.
\newblock Explicit residual descent for 3d human pose estimation from 2d joint
  locations.
\newblock In {\em BMVC}, 2020.

\bibitem{chen20173d}
Ching-Hang Chen and Deva Ramanan.
\newblock 3d human pose estimation= 2d pose estimation+ matching.
\newblock In {\em CVPR}, 2017.

\bibitem{iqbal2018dual}
Umar Iqbal, Andreas Doering, Hashim Yasin, Bj{\"o}rn Kr{\"u}ger, Andreas Weber,
  and Juergen Gall.
\newblock A dual-source approach for 3d human pose estimation from single
  images.
\newblock {\em Computer Vision and Image Understanding}, 172:37--49, 2018.

\bibitem{pons2014posebits}
Gerard Pons-Moll, David~J Fleet, and Bodo Rosenhahn.
\newblock Posebits for monocular human pose estimation.
\newblock In {\em CVPR}, 2014.

\bibitem{choi2020pose2mesh}
Hongsuk Choi, Gyeongsik Moon, and Kyoung~Mu Lee.
\newblock Pose2mesh: Graph convolutional network for 3d human pose and mesh
  recovery from a 2d human pose.
\newblock ECCV, 2020.

\bibitem{mehta2017monocular}
Dushyant Mehta, Helge Rhodin, Dan Casas, Pascal Fua, Oleksandr Sotnychenko,
  Weipeng Xu, and Christian Theobalt.
\newblock Monocular 3d human pose estimation in the wild using improved cnn
  supervision.
\newblock In {\em 3DV}, 2017.

\bibitem{mehta2017vnect}
Dushyant Mehta, Srinath Sridhar, Oleksandr Sotnychenko, Helge Rhodin, Mohammad
  Shafiei, Hans-Peter Seidel, Weipeng Xu, Dan Casas, and Christian Theobalt.
\newblock Vnect: Real-time 3d human pose estimation with a single rgb camera.
\newblock {\em ACM Transactions on Graphics}, 36(4):1--14, 2017.

\end{thebibliography}
}

\end{document}


\title{Supplementary Material - Everybody Is Unique: Towards Unbiased Human Mesh Recovery}
\maketitle

\section{Sample Data}
\begin{figure*}[h!]
\centering
\includegraphics[scale=0.4]{figs/sample.png}
\caption{Sample images from (a) SSP-3D \cite{sengupta2020synthetic} and (b) LargeSize.}
\label{fig:sample}
\end{figure*}

Figure \ref{fig:sample} shows sample images from SSP-3D dataset \cite{sengupta2020synthetic} and our LargeSize dataset.

\section{Problems with EFT}
\begin{figure}[h!]
\centering
\includegraphics[scale=0.25]{figs/EFTbad2.png}
\caption{The comparison of mesh fitting results with EFT \cite{joo2020exemplar} and our proposed OMR.}
\label{fig:EFTbad}
\end{figure}

Figure \ref{fig:EFTbad} shows some fitting results with EFT \cite{joo2020exemplar}, which illustrate the problmes mentioned in Section 3.2, i.e. the risk of overfitting the 2D cost function and biased estimation for obese data. However, our proposed OMR addresses these limitations.

\section{Implementation Details}
\subsection{Mesh Vertices to Segmentation Masks}
Given the mesh vertices, we use a differentiable renderer SoftRas \cite{liu2019soft}, which fuses the probabilistic contributions of all mesh triangles with respect to the rendered pixels, and a predefined texture map to generate the binary body part segmentation masks $\bm{S} \in \mathbb{R}^{H\times W \times D}$, where $H$, $W$ and $D$ are the height, width and the number of body part, respectively.

\subsection{Loss Items}
We use the Geman-McClure error function \cite{ganan1985bayesian} to measure the 2D re-projection loss $L_\text{2D}$ for both $P$-iteration and $Q$-iteration steps:
\begin{equation}
    L_\text{2D}(\hat{\mathbf{x}},\mathbf{x}) = \frac{\sigma^2 * (\hat{\mathbf{x}}-\mathbf{x})^2}{\sigma^2 + (\hat{\mathbf{x}}-\mathbf{x})^2},
    \label{eq:l2d}
\end{equation}
where $\sigma=100$, and $\hat{\mathbf{x}}$ and $\mathbf{x}$ are the predicted 2D joints and their corresponding ground truth. The pose prior $L_\theta(\theta)$ is implemented following prior work \cite{pavlakos2019expressive} as:
\begin{equation}
    L_{\theta}(\theta) = ||Z(\theta)||^2_2,
    \label{eq:ltheta}
\end{equation}
where $Z(\theta)$ is the latent representation learned by a variational autoencoder.

\subsection{Optimization}
We use the Adam optimizer \cite{kingma2014adam} for both $P$-iteration and $Q$-iteration steps. The learning rate for the $Q$-iteration step is set to 1e-3, whereas the learning rate of $P$-iteration step is set to 1e-6. The number of iterations for each single $P$-iteration and $Q$-iteration steps is 20. All our implementation is in PyTorch \cite{NEURIPS2019_9015}.  

\subsection{SSP-3D \cite{sengupta2020synthetic} Validation Set}
To select data with extreme shape parameters, we measure the PVE-T between the mean shape parameters and those of each sample in SSP-3D. The data whose PVE-T is no less than 22.5 mm will be included into the validation set.

\section{Experimental Results}
\subsection{Quantitative Results}
Table \ref{tab:ssphkmr} shows results on the SSP-3D dataset where we use HKMR \cite{georgakis2020hierarchical} as the base model. Similarly, We can observe the proposed $L_{\text{shape}}$ consistently reduces both PA-MPJPE and PVE-T errors across all the methods, and the proposed OMR
outperforms both SMPLifty \cite{bogo2016keep} and EFT \cite{joo2020exemplar}.
\begin{table*}[h!]
\begin{center}
\scalebox{0.9}{
\begin{tabular}{ccc}
\toprule
SSP-3D              & MPJPE-PA (mm)     & PVE-T (mm) \\
\midrule
HKMR                    & 57.53     & 34.54  \\
\midrule
SMPLify w/o $L_{\text{shape}}$   & 55.02     & 28.92  \\
SMPLify+$L_{\text{shape}}$           & 53.31     & 27.20  \\
\midrule
EFT w/o $L_{\text{shape}}$        & 55.29     & 30.17  \\
EFT+$L_{\text{shape}}$               & 54.88     & 29.59  \\
\midrule
OMR w/o $L_{\text{shape}}$  & 52.67 & 29.20 \\
OMR+$L_{\text{shape}}$             & \textbf{49.77}     & \textbf{18.72}  \\
\bottomrule
\end{tabular}
~~
\begin{tabular}{ccccc}
\toprule
SSP-3D (PVE-T)          & Torso    & Legs    & Arms    & Head \\
\midrule
HKMR             & 52.10    & 25.04   & 36.93   & 31.19 \\
\midrule
SMPLify w/o $L_{\text{shape}}$    & 47.05    & 19.41   & 32.50   & 23.70 \\
SMPLify+$L_{\text{shape}}$    & 43.99    & 18.68   & 30.30   & 22.02 \\
\midrule
EFT w/o $L_{\text{shape}}$         & 49.10    & 20.01   & 34.22   & 24.86 \\
EFT+$L_{\text{shape}}$         & 48.17    & 19.59   & 33.12   & 24.94 \\
\midrule
OMR w/o $L_{\text{shape}}$  & 47.68 & 19.25 & 32.30 & 25.00 \\
OMR+$L_{\text{shape}}$             & \textbf{24.46}     & \textbf{16.04}    & \textbf{19.84}  & \textbf{16.38}\\
\bottomrule
\end{tabular}}
\end{center}
\caption{SMPLify \cite{bogo2016keep} vs. EFT \cite{joo2020exemplar} vs. proposed OMR on SSP-3D using HKMR \cite{georgakis2020hierarchical} as the base model.}
\label{tab:ssphkmr}
\end{table*}

Table \ref{tab:acrossModelCMR} shows results on Human3.6M dataset for the generalized model fitting evaluation, where CMR \cite{kolotouros2019convolutional} is the base model. The proposed OMR is still able to have a steady decrease in both MPJPE and PA-MPJPE, which reaches to the lowest error.
\begin{table}[h!]
\begin{center}
\scalebox{1}{
\begin{tabular}{ccc}
\toprule
Human3.6M              & MPJPE     & PA-MPJPE \\
\midrule
CMR \cite{kolotouros2019convolutional}                   & 76.04     & 50.46  \\
\midrule
SMPLify - 20           & 84.44     & 57.41  \\
SMPLify - 100          & 101.92    & 63.38  \\
\midrule
EFT - 20               & 70.32     & 47.43  \\
EFT - 100              & 74.62     & 46.77  \\
\midrule
OMR (1P1Q)            & 70.03     & 47.10  \\
OMR (5P4Q)            & \textbf{68.89}     & \textbf{44.56}  \\
\bottomrule
\end{tabular}
}
\end{center}
\caption{SMPLify vs. EFT vs. OMR on Human3.6M using the CMR base model. All numbers in mm.}
\label{tab:acrossModelCMR}
\end{table}

Table \ref{tab:RetrainingHMRCMR} shows the performance improvement obtained by using OMR-generated parameters for the training of HMR \cite{kanazawa2018end} and CMR \cite{kolotouros2019convolutional}.

Finally, in Table \ref{tab:RetrainingLargeSize}, we show results of retraining HMR, CMR, SPIN, and HKMR on our LargeSize data, where one can note substantial performance improvements with the proposed OMR. 
\begin{table*}[h!]
  \begin{center}
  \scalebox{0.88}{
    \begin{tabular}{ccccc}
    \toprule
     SSP-3D    &   MPJPE  &  PA-MPJPE  &    PVE-T  &  mIoU  \\
    \midrule
     HMR \cite{kanazawa2018end}  & 102.34 & 67.91  & 31.41  & 0.6477 \\
     HMR - OMR                   & \textbf{99.54} & \textbf{59.83}  & \textbf{30.87}  & \textbf{0.6599} \\
     \midrule
     CMR \cite{kolotouros2019convolutional}  & 135.51  & 67.11  & 730.62  & 0.6651 \\
     CMR - OMR                               & \textbf{93.61}  & \textbf{56.19}  & \textbf{30.08}  & \textbf{0.6862} \\
    \bottomrule
    \end{tabular}}
    ~~
    \scalebox{0.8}{
    \begin{tabular}{ccccc}
    \toprule
      Human3.6M   & \multicolumn{2}{c}{Protocol \#1}  &  \multicolumn{2}{c}{Protocol \#2}  \\
    \cline{2-5}
         &   MPJPE  &  PA-MPJPE  &    MPJPE  &  PA-MPJPE  \\
    \midrule
     HMR \cite{kanazawa2018end}                         & 87.97 & 58.1  & 88.0  & 56.8 \\
     HMR - OMR               & \textbf{77.73} & \textbf{56.1}  & \textbf{74.2}  & \textbf{53.8} \\
     \midrule
     CMR \cite{kolotouros2019convolutional}             & 74.7  & 51.9  & 71.9  & 50.1 \\
     CMR - OMR    & \textbf{67.0}  & \textbf{47.9}  & \textbf{64.7}  & \textbf{45.7} \\
    \bottomrule
    \end{tabular}}
  \end{center}
  \caption{Improving baseline models by retraining with annotations generated by our method.}
      \label{tab:RetrainingHMRCMR}
\end{table*}

\subsection{Qualitative Results}
In Figure \ref{fig:meshFitting}, we show some representative mesh fitting results using the proposed OMR on both the generic standard benchmark dataset and our LargeSize dataset. Figure \ref{fig:3dpw} and \ref{fig:retrainLS} show the improved mesh estimations with HKMR \cite{georgakis2020hierarchical} and SPIN \cite{kolotouros2019learning} on 3DPW \cite{vonMarcard2018} and LargeSize when compared to their corresponding baseline versions. 
\begin{figure*}[h!]
\centering
\includegraphics[scale=0.6]{figs/fit.png}
\caption{OMR mesh fits on (a) non-obese standard benchmark data and (b) LargeSize data.}
\label{fig:meshFitting}
\end{figure*}

\begin{figure*}[h!]
\centering
\includegraphics[scale=0.43]{figs/3dpw.png}
\caption{Mesh estimation results on 3DPW \cite{vonMarcard2018} of HKMR and SPIN before and after retraining with OMR-generated parameters.}
\label{fig:3dpw}
\end{figure*}

\begin{figure*}[h!]
\centering
\includegraphics[scale=0.43]{figs/supp-retrainShape.png}
\caption{Mesh estimation results on LargeSize of HKMR and SPIN before and after retraining with OMR-generated parameters.}
\label{fig:retrainLS}
\end{figure*}

\begin{table}[h!]
  \begin{center}
  \scalebox{0.9}{
     \begin{tabular}{cccccc}
    \toprule
      & \multicolumn{2}{c}{FB Seg.}  &  \multicolumn{2}{c}{Part Seg.} & \multirow{2}{1.cm}{\centering PVE-T}\\
    \cline{2-5}
         &   acc.  &  f1  &   acc. &  f1  \\
    \midrule
     HMR \cite{kanazawa2018end}  & 93.89 & 89.35  & 91.97  & 71.37 & 36.87 \\
     HMR - OMR                   & \textbf{94.71} & \textbf{91.01}  & \textbf{92.99}  & \textbf{76.21}  &  \textbf{20.24}\\
     \midrule
     CMR \cite{kolotouros2019convolutional}  & 93.93 & 89.84  & 91.83  & 72.26 & 24.93 \\
     CMR - OMR                               & \textbf{94.45}  & \textbf{90.42}  & \textbf{92.90}  & \textbf{76.30} & \textbf{22.05}\\
     \midrule
     SPIN \cite{kolotouros2019learning}  & 93.76 & 89.01  & 92.20 & 73.25 & 29.24 \\
     SPIN - SMPLify                      & 94.93 & 91.37  & 93.41 & 77.94 & 20.42 \\
     SPIN - EFT                          & 94.73 & 90.89  & 93.27 & 77.91 & 21.55 \\
     SPIN - OMR                          & \textbf{95.78} & \textbf{93.13} & \textbf{94.32} & \textbf{81.01} & \textbf{15.20} \\
     \midrule
     HKMR \cite{georgakis2020hierarchical}  & 94.42 & 90.35  & 92.82 & 75.77 & 30.73 \\
     HKMR - SMPLify                         & 94.76 & 90.98  & 93.23 & 76.97 & 20.30 \\
     HKMR - EFT                             & 94.57 & 90.55  & 93.10 & 76.92 & 21.67 \\
     HKMR - OMR                             & \textbf{95.82} & \textbf{93.00}  & \textbf{94.29} & \textbf{80.57} & \textbf{13.98}\\
    \bottomrule
    \end{tabular}}
  \end{center}\caption{Improving baseline models by retraining with annotations generated by our method on Our LargeSize dataset.}
      \label{tab:RetrainingLargeSize}
\end{table}

{\small
\bibliographystyle{ieee_fullname}
\bibliography{egbib}
}